\pdfoutput=1
\documentclass[a4paper,twoside]{article}

\usepackage{epsfig}
\usepackage{caption}
\usepackage{subcaption}
\usepackage{graphicx}
\usepackage{calc}
\usepackage{amssymb}
\usepackage{amstext}
\usepackage{amsmath}
\usepackage{amsthm}
\usepackage{multicol}
\usepackage{pslatex}
\usepackage{apalike}
\usepackage{algorithm2e}
\usepackage[bottom]{footmisc}
\usepackage{hyperref}
\usepackage{listings}
\usepackage{booktabs}
\usepackage{multirow}
\usepackage{array}   
\usepackage[table]{xcolor}
\usepackage{SCITEPRESS}     

\begin{document}

\title{Predicting LLM Correctness in Prosthodontics Using Metadata and Hallucination Signals}
\author{\authorname{Lucky Susanto\sup{1}\orcidAuthor{0009-0003-3494-9249},
Anasta Pranawiyana\sup{2},
Cortino Sukotjo\sup{3}\orcidAuthor{0000-0002-2171-004X},
Soni Prasad\sup{4}\orcidAuthor{0000-0002-4297-5012}, 
Derry Wijaya\sup{1,5}\orcidAuthor{0000-0002-0848-4703}
}
\affiliation{\sup{1}Department of Data Science, Monash University Indonesia, Tangerang, Indonesia}
\affiliation{\sup{2}Independent Researcher}
\affiliation{\sup{3}Department of Prosthodontics, University of Pittsburgh, Pittsburgh, Pennsylvania}
\affiliation{\sup{4}Department of Restorative Sciences, University of North Carolina Adams School of Dentistry, Chapel Hill, North Carolina}
\affiliation{\sup{5}Department of Computer Science, Boston University, Boston, Massachusetts}
\email{lucky.susanto@monash.edu, adhibaye@gmail.com, COS162@pitt.edu, prasadso@unc.edu, derry.wijaya@monash.edu}
}


\abstract{Large language models (LLMs) are increasingly adopted in high-stakes domains such as healthcare and medical education, where the risk of generating factually incorrect (i.e., hallucinated) information is a major concern. While significant efforts have been made to detect and mitigate such hallucinations, predicting whether an LLM's response is correct remains a critical yet underexplored problem. This study investigates the feasibility of predicting correctness by analyzing a general-purpose model (GPT-4o) and a reasoning-centric model (OSS-120B) on a multiple-choice prosthodontics exam. We utilize metadata and hallucination signals across three distinct prompting strategies to build a correctness predictor for each (model, prompting) pair. Our findings demonstrate that this metadata-based approach can improve accuracy by up to +7.14\% and achieve a precision of 83.12\% over a baseline that assumes all answers are correct. We further show that while actual hallucination is a strong indicator of incorrectness, metadata signals alone are not reliable predictors of hallucination. Finally, we reveal that prompting strategies, despite not affecting overall accuracy, significantly alter the models' internal behaviors and the predictive utility of their metadata. These results present a promising direction for developing reliability signals in LLMs but also highlight that the methods explored in this paper are not yet robust enough for critical, high-stakes deployment.}

\onecolumn \maketitle \normalsize \setcounter{footnote}{0} \vfill

\section{\uppercase{Introduction}}
\label{sec:introduction}
Large language models (LLMs) have seen an increase in adoption, even in high-stakes domains such as healthcare. Despite their capabilities, LLMs are still prone to generating hallucinated content. This is particularly dangerous, as these hallucinations often sound logical yet contain non-factual information \cite{Huang_2025}. This issue is particularly concerning in medical education, where LLMs are increasingly used as learning aids or evaluators. In such settings, hallucinated yet plausible explanations can misinform students or reinforce misconceptions, indirectly affecting clinical competence \cite{alrazac2023medicaleducation}.

Due to the danger of hallucinations, many efforts have been made to detect and mitigate them \cite{manakul2023selfcheckgptzeroresourceblackboxhallucination,kadavath2022languagemodelsmostlyknow}. These works detect hallucinations by utilizing information such as consistency, log probability, entropy distribution, and many more. However, to our knowledge, previous work has not directly attempted to predict model correctness. As previously mentioned, predicting correctness is vital in medical education, where inaccurate or misleading model responses could negatively influence learners' understanding or clinical reasoning.

In this study, we analyze two distinct LLMs on a specialized, non-reasoning task (i.e., a multiple-choice question answering problem) in the field of prosthodontics. We examine how metadata, specifically response consistency and answer token log probability, can be used to predict whether the model will answer the question correctly or not. Moreover, we then compare different prompting strategies and how they affect these two different models. Through this work, we bring the following contributions:
\begin{enumerate}
  \item We demonstrate that it is feasible to predict model correctness by only looking at its consistency and log probabilities, achieving an increased accuracy of up to +7.14\% when compared to simply assuming LLMs are always correct, alongside a maximum precision of 83.12\%.
  \item We show that under an idealized setup, hallucination signals are strong indicators of model correctness. However, this requires a powerful hallucination detection model.
  \item In relation to the previous finding, we also show that consistency and token log probabilities are not predictive of hallucination signals.
  \item Lastly, we found that even though differing prompting strategies have no significant effect on model accuracy, different prompting strategies affect GPT-4o's behavior more than OSS-120B's, as reflected in the predictiveness of metadata.
\end{enumerate}

\section{\uppercase{Related Works}}
\begin{figure*}[t]
  \centering
   {\epsfig{file = 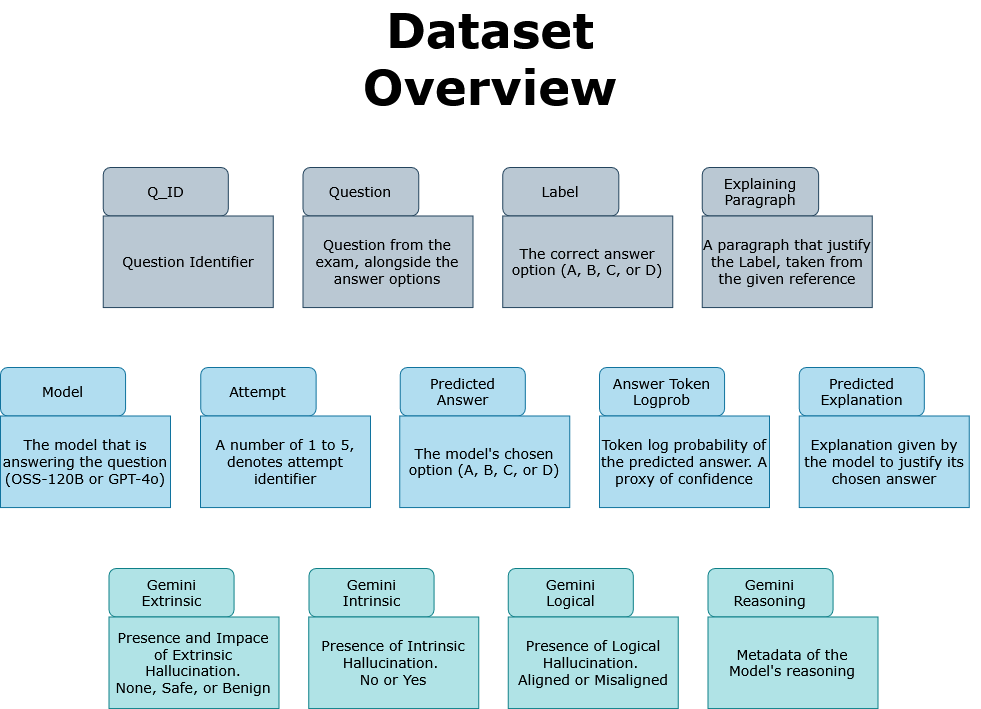, width = 12cm}}
  \caption{Dataset Overview. First row in grey denotes Primary Data Field. Second row in blue denotes Inference Data Field. Third row in cyan denotes LLM-as-a-Judge Data Field.}
  \label{fig:m1_dataset_overview}
\end{figure*}

\subsection{Reasoning LLMs}
Reasoning LLMs are a relatively new technology. Popularized by DeepSeek-R1 \cite{deepseekai2025deepseekr1incentivizingreasoningcapability}, these LLMs show incredible promise in tasks such as math, coding, and other reasoning-heavy tasks \cite{comanici2025gemini25,yang2025qwen3technicalreport}. Perhaps due to their novelty, there is only a minimal amount of work analyzing the effect of reasoning on non-reasoning tasks. A recently released paper notes that we should induce reasoning only for specific tasks requiring reasoning, as they found that a general LLM would be more efficient otherwise \cite{boizard2025doesreasoningmattercontrolled}.

\subsection{Prompting Strategies}
After the rise of LLMs, a particular prompting strategy rose above the rest: Chain-of-Thought prompting. Shortened to CoT, this prompting strategy has been shown to induce reasoning capabilities in LLMs \cite{wei2023chainofthoughtpromptingelicitsreasoning}. In fact, reasoning LLMs such as DeepSeek-R1 are trained through post-training on reasoning-oriented datasets containing CoT exemplars, designed to incentivize step-by-step reasoning. 

However, recent work has cast doubt on this approach, as it has been shown that CoT prompting can be unfaithful \cite{turpin2023languagemodelsdontsay} and that the reasoning induced by CoT is potentially a mirage \cite{zhao2025chainofthoughtreasoningllmsmirage}. Specifically, it seems that CoT is used by the model to match the pattern of reasoning, not to perform reasoning itself. In contrast, Answer-then-Explain (post-hoc explanation) is less popular, though there exist works that utilize it to improve model performance \cite{krishna2023posthocexplanationslanguage}. 

Lastly, it has been reported that despite differing prompting strategies having the same average faithfulness score, they can actually have a high degree of disagreement in the explanations they generate \cite{huang2023largelanguagemodelsexplain}. This shows that different prompting strategies affect the behavior of how LLMs produce their answers. However, this work has only been done on general LLMs, leaving room to explore this for reasoning LLMs.

\subsection{LLM-as-a-Judge}
Due to the impressive capabilities of modern LLMs, a new paradigm has recently emerged. Known as "LLM-as-a-judge" \cite{li2025generationjudgmentopportunitieschallenges}, this method utilizes an LLM to act as a judge that rates the output of another model based on specific metrics. Of course, it is possible that the judge itself is prone to error, thus choosing a reliable model is a must. Even when a reliable model is chosen, it is imperative to understand that LLMs are not perfect.

\section{\uppercase{Methodology}}

\subsection{Dataset}
We use the 2025 National Prosthodontics Resident Exam questions in this work. The exam originally consists of 150 multiple-choice questions. We remove questions that use books as the source of reference due to the difficulty of extracting ground truth explanations. Furthermore, we exclude questions that include images. After filtering, we end up with a final dataset of 98 questions.

To obtain the ground truth explanation, we use GPT-4o to extract an explanatory paragraph with respect to a given question and answer. The extracted paragraph is then evaluated by two experts in the field of prosthodontics, ensuring its quality. Any incorrectly extracted paragraphs were then manually fixed by the authors. Refer to Figure \ref{fig:m1_dataset_overview}'s Primary Data Field for an overview.

\subsection{Models}
We evaluate two models in this work:
\begin{enumerate}
    \item \textbf{GPT-4o} \cite{openai2024gpt4ocard}\textbf{:} A proprietary generalist LLM
    \item \textbf{OSS-120B} \cite{openai2025gptoss120bgptoss20bmodel}\textbf{:} An open-source reasoning LLM
\end{enumerate}
We prompt each model five times for each promting strategy. Refer to Figure~\ref{fig:m1_dataset_overview}'s Inference Data Field for an overview.

\subsection{Prompting Strategy}
We use three types of prompting in this work:
\begin{enumerate}
    \item \textbf{Answer only (A):} Provided a question, the model answers directly.
    \item \textbf{Answer then Explain (AE):} Provided a question, the model answers and then provides a justification.
    \item \textbf{Explain then Answer (EA):} Commonly known as Chain-of-Thought; provided a question, the model reasons to reach an answer.
\end{enumerate}

We use the following system prompt to enforce a consistent answer format for both models:
\begin{lstlisting}[
    basicstyle=\sffamily\small,
    columns=fullflexible,       
    breaklines=true,
    frame=tb,         
    mathescape=false
]
Instruction: You must answer this question using the exact template below
Response:
Answer: Only the selected option (e.g., A [without anything else])
Explanation: Your explanation
\end{lstlisting}
To change the prompting strategy, we modify the system prompt as follows:
\begin{enumerate}
  \item Leave it unchanged for \textbf{Answer then Explain}.
  \item Move the "Explanation" line above the "Answer" line for \textbf{Explain then Answer}.
  \item Delete the "Explanation" line for the \textbf{Answer only} strategy.
\end{enumerate}

\subsection{Evaluation}
For each question in the dataset, we prompt each model five times. We then take the majority answer as the model's final answer.

\subsection{Hallucinations}
We use \textbf{LLM-as-a-judge} \cite{li2025generationjudgmentopportunitieschallenges}, with Gemini-2.5-flash \cite{comanici2025gemini25} as a judge. This model is chosen due to its capabilities, with a reported hallucination rate of 1.3\% on the Vectara hallucination benchmark \cite{hhem-2.1-open}\footnote{As reported on the
\href{https://huggingface.co/spaces/vectara/leaderboard}{Vectara LLM Hallucination Leaderboard}.}. The prompt is available in Appendix \ref{app:1_haluprompt}. Note that the Vectara hallucination benchmark is not specific to the prosthodontics domain and thus this number may be an overestimation of the model's capabilities.

In this work, we focus on three types of hallucinations:
\begin{enumerate}
    \item \textbf{Intrinsic Hallucination:} The explanation provided by the model directly contradicts the reference explanation.
    \item \textbf{Extrinsic Hallucination:} The explanation provided by the model contains unverifiable information with respect to the reference explanation.
    \item \textbf{Logical Hallucination:} The explanation provided by the model does not support its final answer.
\end{enumerate}

We further refine the definition for \textbf{Extrinsic Hallucination}:
\begin{enumerate}
    \item \textbf{Safe:} There is no unverifiable information from the model's generated explanation.
    \item \textbf{Benign:} The unverifiable information generated by the model is superfluous and the core premise does not differ from the reference explanation.
    \item \textbf{Harmful:} The unverifiable information generated by the model has a core premise that differs from the reference explanation.
\end{enumerate}
Examples illustrating the difference between classes of extrinsic hallucination are provided in Appendix \ref{app:2_extrinsic}. In this work, we group extrinsic hallucinations into two categories: harmful and non-harmful (i.e., safe or benign). We made this choice because our setup limits the reference explanation to a single paragraph. However, this simplification introduces a limitation: a model's explanation may be factual yet unverifiable within the relatively short reference. This results in a higher occurrence of benign extrinsic hallucinations that would have been deemed safe if the reference explanation were more complete. Refer to Figure \ref{fig:m1_dataset_overview}'s LLM-as-a-Judge Data Field for an overview.

\subsection{Metadata}
This work chooses two types of metadata as the focus:
\begin{enumerate}
    \item \textbf{Consistency}: How often each model outputs its most common answer across five attempts.
    \item \textbf{Answer token log probability}: A proxy for the model's confidence when deciding which choice it should answer with (i.e., X in "Answer: [X]"). A log probability of 0 denotes absolute confidence.
\end{enumerate}

\section{\uppercase{Exploratory Analysis}}
\subsection{Model Performance}
\begin{figure}[!h]
  \centering
   {\epsfig{file = 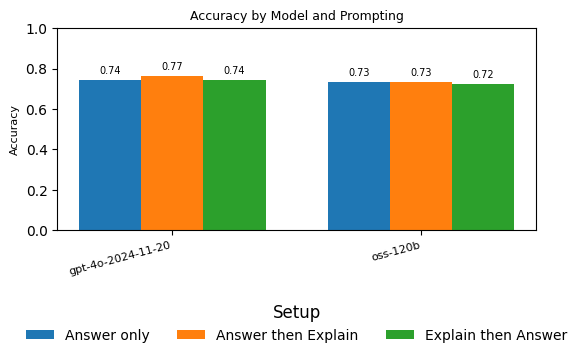, width = 7.5cm}}
  \caption{Different prompting strategies have a non-significant effect on model accuracy.}
  \label{fig:1_acc}
\end{figure}
Despite previous works reporting Reasoning LLMs' higher capabilities in domains such as math, coding, and other reasoning-heavy tasks, we found results indicating that this ability does not help in a non-reasoning task, supporting results from previous works \cite{boizard2025doesreasoningmattercontrolled}. 

\subsection{Consistency}
\begin{figure}[!b]
  \centering
   {\epsfig{file = 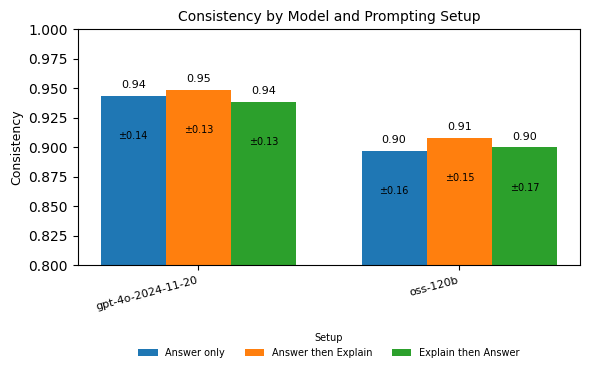, width = 7.5cm}}
  \caption{Different prompting strategies have a non-significant effect on model consistency.}
  \label{fig:2_consistency}
\end{figure}
We find that GPT-4o is relatively more consistent across the board when compared to OSS-120B. However, differences in prompting methods do not significantly affect consistency.

\subsection{Disagreement}
\begin{figure}[!h]
  \vspace{-0.2cm}
  \centering
   {\epsfig{file = 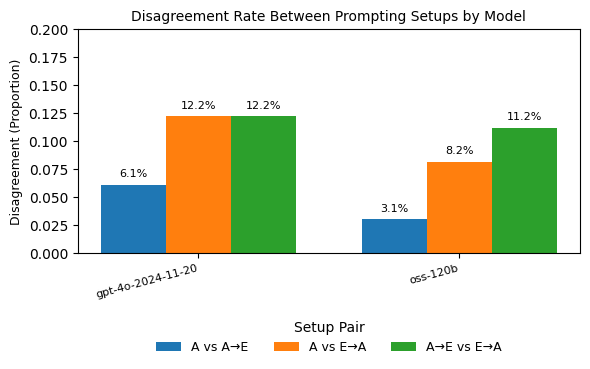, width = 7.5cm}}
  \caption{Despite similar performance, each pair of prompting strategies shows some degree of disagreement.}
  \label{fig:3_disagreement}
\end{figure}
Despite having similar accuracy, each prompting strategy impacts what each model eventually responds with.  

\subsection{Correlation}
\begin{figure}[!b]
  \centering
   {\epsfig{file = 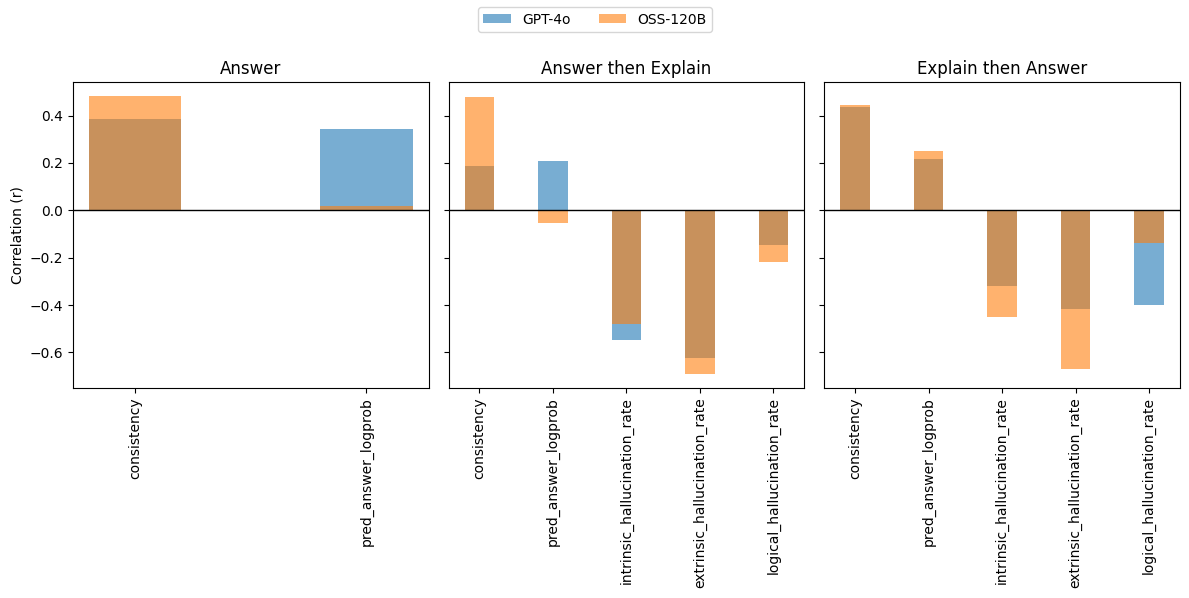, width = 7cm}}
  \caption{Correlation of each metadata and hallucination rate with respect to each model's correctness. We group safe and benign extrinsic hallucinations into one group (non-harmful extrinsic hallucination) and compare it to harmful extrinsic hallucination.}
  \label{fig:4_correlation}
\end{figure}
The correlation plot shows that consistency has a positive correlation with the model's correctness for both GPT-4o and OSS-120B. However, except in the case of \textit{Explain then Answer}, answer token log probability has near-zero correlation for OSS-120B. We note that OSS-120B is extremely confident in most cases, with an answer token log probability near zero (indicating absolute confidence). 

We define hallucination rate as the average occurrence of each hallucination for a specific question when prompted five times. As previously mentioned, we group safe and benign extrinsic hallucinations into one group, treating them as if no extrinsic hallucination occurred, with harmful extrinsic hallucination being the other case.

\section{\uppercase{Result: Predicting Correctness}}
\begin{table*}[t]
\centering
\scriptsize
\caption{Performance comparison of GPT-4o and OSS-120B across prompting strategies. \textit{Gray rows (Oracle)} denote upper-bound potential, not comparable results. Value inside brackets indicates improvement compared to baseline (assuming that all output made by an LLM is always correct). Under the baseline scenario, recall achieve a perfect score of 1. REG (regression) and CLS (classification) refers to the addition of predicted hallucination signals. Oracle refers to the addition of actual hallucination signals, where \textit{Rate} denotes the average occurence of each hallucination and \textit{Maj} denotes whether or not each hallucination occurs at least three times when the question is prompted five times.}
\label{tab:1-main}
\begin{tabular}{cllccccc}
\toprule
\# & \textbf{Model} & \textbf{Prompting} & \textbf{Accuracy} & \textbf{Precision} & \textbf{Recall} & \textbf{F1} & \textbf{ROC-AUC} \\
\midrule
1  & \textbf{GPT-4o} & Answer Only & 78.57\% (+4.08\%) & 78.89\% (+4.40\%) & 97.26\% (-2.74\%) & 87.12\% (+1.74\%) & 0.573 \\
2  & \textbf{GPT-4o} & Answer then Explain & 74.49\% (-2.04\%) & 76.60\% (+0.07\%) & 96.00\% (-4.00\%) & 85.21\% (-1.50\%) & 0.484 \\
3  & \textbf{GPT-4o} & \hspace*{1em}+ REG & 77.55\% (+1.02\%) & 78.49\% (+1.96\%) & 97.33\% (-2.67\%) & 86.90\% (+0.19\%) & 0.763 \\
4  & \textbf{GPT-4o} & \hspace*{1em}+ CLS & 78.57\% (+2.04\%) & 79.35\% (+2.82\%) & 97.33\% (-2.67\%) & 87.43\% (+0.72\%) & 0.710 \\
\rowcolor{gray!10}
5  & \textbf{GPT-4o} & \hspace*{1em}+ (Oracle) Rate & 85.71\% (+9.18\%) & 89.61\% (+13.08\%) & 92.00\% (-8.00\%) & 90.79\% (+4.08\%) & 0.897 \\
\rowcolor{gray!10}
6  & \textbf{GPT-4o} & \hspace*{1em}+ (Oracle) Maj & 84.69\% (+8.16\%) & 87.50\% (+10.97\%) & 93.33\% (-6.67\%) & 90.32\% (+3.61\%) & 0.868 \\
7  & \textbf{GPT-4o} & Explain then Answer & 79.59\% (+5.10\%) & 80.46\% (+5.97\%) & 95.89\% (-4.11\%) & 87.50\% (+2.12\%) & 0.822 \\
8  & \textbf{GPT-4o} & \hspace*{1em}+ REG & 80.61\% (+6.12\%) & 81.40\% (+6.91\%) & 95.89\% (-4.11\%) & 88.05\% (+2.67\%) & 0.815 \\
9  & \textbf{GPT-4o} & \hspace*{1em}+ CLS & 80.61\% (+6.12\%) & 82.14\% (+7.65\%) & 94.52\% (-5.48\%) & 87.90\% (+2.52\%) & 0.803 \\
\rowcolor{gray!10}
10 & \textbf{GPT-4o} & \hspace*{1em}+ (Oracle) Rate & 78.57\% (+4.08\%) & 83.33\% (+8.84\%) & 89.04\% (-10.96\%) & 86.09\% (+0.71\%) & 0.889 \\
\rowcolor{gray!10}
11 & \textbf{GPT-4o} & \hspace*{1em}+ (Oracle) Maj & 82.65\% (+8.16\%) & 85.90\% (+11.41\%) & 91.78\% (-8.22\%) & 88.74\% (+3.36\%) & 0.890 \\
\midrule
12 & \textbf{OSS-120B} & Answer Only & 77.55\% (+4.08\%) & 81.25\% (+7.78\%) & 90.28\% (-9.72\%) & 85.53\% (+0.82\%) & 0.643 \\
13 & \textbf{OSS-120B} & Answer then Explain & 78.57\% (+5.10\%) & 80.72\% (+7.25\%) & 93.06\% (-6.94\%) & 86.45\% (+1.74\%) & 0.591 \\
14 & \textbf{OSS-120B} & \hspace*{1em}+ REG & 79.59\% (+6.12\%) & 80.95\% (+7.48\%) & 94.44\% (-5.56\%) & 87.18\% (+2.47\%) & 0.594 \\
15 & \textbf{OSS-120B} & \hspace*{1em}+ CLS & 79.59\% (+6.12\%) & 80.95\% (+7.48\%) & 94.44\% (-5.56\%) & 87.18\% (+2.47\%) & 0.613 \\
\rowcolor{gray!10}
16 & \textbf{OSS-120B} & \hspace*{1em}+ (Oracle) Rate & 84.69\% (+11.22\%) & 89.04\% (+15.57\%) & 90.28\% (-9.72\%) & 89.66\% (+4.95\%) & 0.929 \\
\rowcolor{gray!10}
17 & \textbf{OSS-120B} & \hspace*{1em}+ (Oracle) Maj & 89.80\% (+16.33\%) & 91.89\% (+18.42\%) & 94.44\% (-5.56\%) & 93.15\% (+8.44\%) & 0.939 \\
18 & \textbf{OSS-120B} & Explain then Answer & 79.59\% (+7.14\%) & 83.12\% (+10.67\%) & 90.14\% (-9.86\%) & 86.49\% (+2.47\%) & 0.736 \\
19 & \textbf{OSS-120B} & \hspace*{1em}+ REG & 79.59\% (+7.14\%) & 83.12\% (+10.67\%) & 90.14\% (-9.86\%) & 86.49\% (+2.47\%) & 0.740 \\
20 & \textbf{OSS-120B} & \hspace*{1em}+ CLS & 79.59\% (+7.14\%) & 83.12\% (+10.67\%) & 90.14\% (-9.86\%) & 86.49\% (+2.47\%) & 0.733 \\
\rowcolor{gray!10}
21 & \textbf{OSS-120B} & \hspace*{1em}+ (Oracle) Rate & 88.78\% (+16.33\%) & 92.86\% (+20.41\%) & 91.55\% (-8.45\%) & 92.20\% (+8.18\%) & 0.890 \\
\rowcolor{gray!10}
22 & \textbf{OSS-120B} & \hspace*{1em}+ (Oracle) Maj & 89.80\% (+17.35\%) & 92.96\% (+20.51\%) & 92.96\% (-7.04\%) & 92.96\% (+8.94\%) & 0.894 \\
\bottomrule
\end{tabular}
\end{table*}

This section addresses the main research question: "Can we predict correctness based on available metadata?" To answer this question, we use leave-one-out cross validation (LOOCV)\footnote{\url{https://scikit-learn.org/stable/modules/generated/sklearn.model_selection.LeaveOneOut.html}} due to our limited number of data points. Table \ref{tab:1-main} reports the results for all experimental setups and a visualizes the experiment flow can be found in the Appendix, Figure \ref{fig:app_workflow}.

\subsection{Using Metadata}
We initialize this experiment using only two metadata features: consistency and the log probability of the answer token. The results are shown in Table \ref{tab:1-main}, specifically in rows 1, 2, 7, 12, 13, and 18. Although previous analyses indicate that prompting strategy has no statistically significant effect on model accuracy, the table shows that the \textit{Explain then Answer} strategy produces metadata that is more predictive of correctness. For GPT-4o, \textit{Explain then Answer} yields a +5.12\% improvement in accuracy over the baseline. For OSS-120B, while \textit{Explain then Answer} provides a +7.14\% increase, other prompting strategies perform comparably, suggesting that prompting order has a smaller impact on metadata predictiveness for this model.

\subsection{Adding Predicted Hallucination Labels}
Obtaining labels for extrinsic and intrinsic hallucinations necessitates the availability of a golden reference, which is typically unattainable in practical scenarios. Consequently, we employ metadata as input to train two predictor models: (1) a classifier that predicts a binary label (i.e., whether a hallucination occurs) and (2) a regressor that estimates the hallucination rate (i.e., the average proportion of hallucinated content across five attempts). Under the present configuration, the consistency and answer token log probability \textbf{are not predictive} of hallucination occurrence. The outcomes of these two models are provided in Appendix \ref{app:3-fd}, Table \ref{tab:app-halupredictor}.

Although Table \ref{tab:1-main} indicates that incorporating predicted hallucination labels marginally improves the correctness predictor model, we attribute this effect to random noise. In most cases, the inclusion of hallucination labels results in either negligible improvements (see rows 7–9 and 13–15) or no improvement at all (see rows 18–20). The apparent performance gains observed in rows 2–4 are likewise interpreted as artifacts of random fluctuations, compounded by the weak baseline performance reported in row 2.

\subsection{Oracle Setting}
The oracle setting represents the upper-bound performance of the correctness predictor when actual hallucination labels are provided. As shown in the gray rows of Table \ref{tab:1-main}, there remains substantial room for improvement. In this idealized setup, GPT-4o with the \textit{Answer then Explain} strategy provides the highest predictability of correctness. For OSS-120B, while the potential for improvement remains large, the upper bounds for \textit{Answer then Explain} and \textit{Explain then Answer} are identical.

\begin{figure}[!t]
  \centering
   {\epsfig{file = 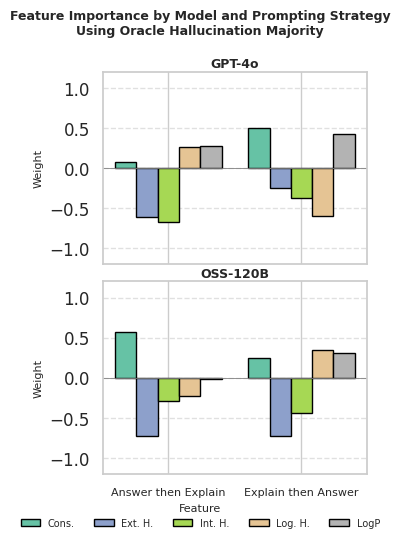, width = 7cm}}
  \caption{Feature importance weights of the correctness predictor when provided with metadata and oracle \textbf{hallucination majority} signal. \textbf{Cons.} = consistency, \textbf{Ext. H.} = extrinsic hallucination, textbf{Int. H.} = intrinsic hallucination,\textbf{Log. H.} = logical hallucination, \textbf{LogP} = predicted answer token log probability.}
  \label{fig:5_oracle_maj_featcoef}
\end{figure}

\begin{figure}[!t]
  \centering
   {\epsfig{file = 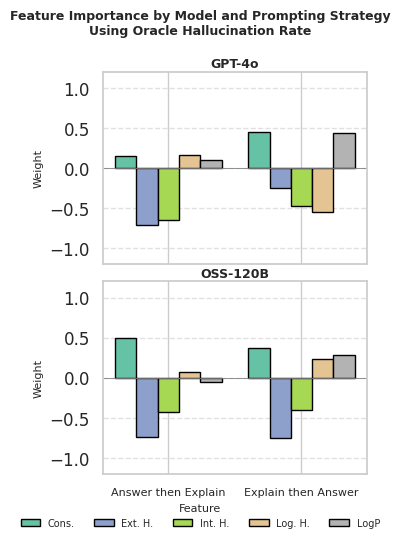, width = 7cm}}
  \caption{Feature importance weights of the correctness predictor when provided with metadata and oracle \textbf{hallucination rate} signal, identical abbreviations as Figure \ref{fig:5_oracle_maj_featcoef}.}
  \label{fig:6_oracle_rate_featcoef}
\end{figure}

Figure \ref{fig:5_oracle_maj_featcoef} and \ref{fig:6_oracle_rate_featcoef} showcase the correctness predictor's feature weights. For GPT-4o, the \textit{Answer then Explain} strategy makes harmful extrinsic and intrinsic hallucinations the strongest predictors of correctness, whereas under \textit{Explain then Answer}, consistency becomes the most important feature. 

For OSS-120B, both prompting strategies highlight consistency and hallucination signals as predictive, though the answer token log probability is only useful under \textit{Explain then Answer}. This observation shows that prompting strategy has an existing effect on the predictiveness of correctness through metadata, despite the same upper-bound performance. We hypothesize that the choice of prompting strategy for OSS-120B still affects the model's internal mechanism, as indicated by the disagreement between strategies, though we leave this question for future work.

\section{\uppercase{Discussion}}
\paragraph{The use of generalist and reasoning LLMs is still too risky in high-stakes domains} Both GPT-4o and OSS-120B achieve only around 75\% and 73\% accuracy on the exam dataset, respectively. This is excessively low for high-stakes domains that require absolute precision. Our work shows that both consistency and answer token log probability are useful for flagging potentially erroneous responses, achieving a maximum precision of 83.12\%, which is still unreliable for high-stakes applications. 

\paragraph{The effect of prompting strategy is still unclear} While we find that different prompting strategies have no effect on model performance for this specific task, they still alter the predictive power of metadata features, which we attribute to differences in the model's internal mechanisms. This is especially surprising for OSS-120B, a reasoning model designed to innately reason. Specifically, we found that \textit{Explain then Answer} may result in different final answers compared to \textit{Answer then Explain}. This is particularly intriguing since the task itself does not require reasoning and is primarily factual.

\paragraph{How different prompting strategies affect LLMs and reasoning LLMs is still understudied} Our results show that even on factual, non-reasoning tasks, prompting strategies can change which features are most predictive of correctness, and can even lead to different final answers for the same model. While GPT-4o shows more pronounced differences between \textit{Answer then Explain} and \textit{Explain then Answer}, OSS-120B exhibits subtler changes in metadata predictiveness despite identical upper-bound performance. These findings suggest that prompting can influence internal decision processes in ways not captured by accuracy alone, highlighting the need for further research into how prompts interact with reasoning models' internal mechanisms across domains and tasks.

\paragraph{Implications for medical education} The exam dataset we use reflects real-world medical knowledge assessments, where accuracy is critical. Our results indicate that even strong LLMs like GPT-4o and OSS-120B achieve only 75–79\% accuracy, corresponding roughly to a B– to B grade—insufficient for high-stakes decisions. While metadata such as consistency and answer log probability can help flag potentially incorrect responses, the maximum precision of 83\% still leaves substantial risk. Even under the oracle setting, accuracy reaches only 89\% (B+), highlighting the gap between current LLM capabilities and safe deployment in medical education contexts. This underscores the importance of cautious, supervised use of LLMs as educational aids rather than as unsupervised decision-makers.

\section{\uppercase{Limitations and Future Works}}
\paragraph{Limited Scope and Small Dataset} The primary limtation of this work lies in the small score (Prosthodontics) and limited dataset size (n=98), which restrict the generalizability and statistical power of our results. The limited scope of prosthodontics is chosen due to the authors field of expertise. Due to the limited dataset size, marginal performance increases should be interpreted with caution. Despite this, the key insight of our work comes from the oracle setting, which demonstrates that correctness is predictable with the help of accurate hallucination detection. We urge future work to replicate this work across other domains and explore more sophisticated and less restrictive hallucination detection methods (i.e., those not requiring a reference explanation). 

\paragraph{No Specialized LLMs} To our knowledge, there are no LLMs specialized in Prosthodontics. Thus, this work uses LLMs trained on broad domains, which may limit domain-specific accuracy and clinical nuance.

\paragraph{Classification of extrinsic hallucination} Our work classifies extrinsic hallucination using a single reference explanation per question. While this simplifies the task, it may misclassify valid alternative explanations as "benign extrinsic hallucinations" when they deviate from the reference. We mitigate this issue by standardizing the annotation into a binary classification: harmful versus non-harmful (i.e., safe or benign) extrinsic hallucinations. This allows us to focus on misleading and factually incorrect explanations.

\section*{\uppercase{Acknowledgements}}
This research was funded by the Ministry of Education, Culture, Research and Technology of the Republic of Indonesia through the Indonesia-US Research Collaboration in Open Digital Technology program, Google Research Scholar Award, and Monash University's Action Lab. The findings and conclusions presented in this publication are those of the authors and do not necessarily reflect the views of the sponsors. 

This work utilizes Gemini-2.5-Pro to help proof-read the writing of the authors and provide suggestions and codes for the visualization shown in many sections of this work. Concretely, for each paragraph we deemed hard to read, we prompt Gemini-2.5-Pro to focus on fixing any grammatical error and increase its readability, which is then re-reviewed by the authors. We also prompt the model to help create the visualization for Figure \ref{fig:4_correlation}, Figure \ref{fig:5_oracle_maj_featcoef}, and Figure \ref{fig:6_oracle_rate_featcoef}.





\bibliographystyle{apalike}
{\small
\bibliography{example}}

\section*{\uppercase{Appendix}}

\subsection*{Hallucination Predictor Model Performance}
\label{app:4_halupredictor}

This appendix shows the performance of the hallucination predictor model, which is available in Table \ref{tab:app-halupredictor}. The metrics used for the regressor are the following:

\begin{enumerate}
  \item \textbf{R$^2$}: The coefficient of the determinator. A value of 0 indicates that the model explains \textbf{none} of the variance in the target variable (the \textbf{Target} column)
  \item \textbf{MAE}: Mean absolute error. A value of 0 indicates perfect model performance.
  \item \textbf{RMSE}: Root mean squared error. A value of 0 indicates perfect model performance.
\end{enumerate}
Note that both the classifier and the regressor perform close to the baseline, suggesting that metadata offers limited predictive power in our current setup. However, this inefficacy is likely driven by severe class imbalance, which may be masking the signal.

\begin{table*}[t]
\centering
\caption{Performance of hallucination predictors using metadata. Left: classification of hallucination class. Right: regression of hallucination rate.}
\label{tab:app-halupredictor}
\begin{tabular}{llccccc|ccc}
\hline
\multicolumn{7}{c|}{\textbf{Classifier}} & 
\multicolumn{3}{c}{\textbf{Regressor}} \\
\hline
Model & Prompting & Target & Baseline Acc & Accuracy & Precision & F1 &
R$^2$ & MAE & RMSE \\
\hline
GPT-4o & A$\rightarrow$E & extrinsic & 0.81 & 0.81 & 0.00 & 0.00 & -0.01 & 0.282 & 0.361 \\
GPT-4o & A$\rightarrow$E & intrinsic & 0.70 & 0.69 & 0.00 & 0.00 & -0.03 & 0.342 & 0.392 \\
GPT-4o & A$\rightarrow$E & logical & 0.98 & 0.98 & 0.00 & 0.00 & -0.02 & 0.051 & 0.117 \\
GPT-4o & E$\rightarrow$A & extrinsic & 0.83 & 0.84 & 0.60 & 0.27 & 0.07 & 0.210 & 0.304 \\
GPT-4o & E$\rightarrow$A & intrinsic & 0.74 & 0.72 & 0.00 & 0.00 & 0.01 & 0.299 & 0.364 \\
GPT-4o & E$\rightarrow$A & logical & 0.90 & 0.90 & 0.00 & 0.00 & -0.00 & 0.178 & 0.236 \\
OSS-120B & A$\rightarrow$E & extrinsic & 0.78 & 0.78 & 0.00 & 0.00 & 0.05 & 0.295 & 0.367 \\
OSS-120B & A$\rightarrow$E & intrinsic & 0.64 & 0.65 & 0.53 & 0.32 & 0.01 & 0.342 & 0.388 \\
OSS-120B & A$\rightarrow$E & logical & 0.98 & 0.98 & 0.00 & 0.00 & 0.05 & 0.071 & 0.115 \\
OSS-120B & E$\rightarrow$A & extrinsic & 0.69 & 0.77 & 0.67 & 0.55 & 0.16 & 0.274 & 0.353 \\
OSS-120B & E$\rightarrow$A & intrinsic & 0.54 & 0.67 & 0.72 & 0.57 & 0.02 & 0.352 & 0.399 \\
OSS-120B & E$\rightarrow$A & logical & 0.98 & 0.98 & 0.00 & 0.00 & -0.05 & 0.080 & 0.124 \\
\hline
\end{tabular}
\label{tab:halluc-predictors}
\end{table*}


\subsection*{Experiment Workflow}
\label{app:experiment_workflow}
\begin{figure*}[!t]
  \centering
   {\epsfig{file = 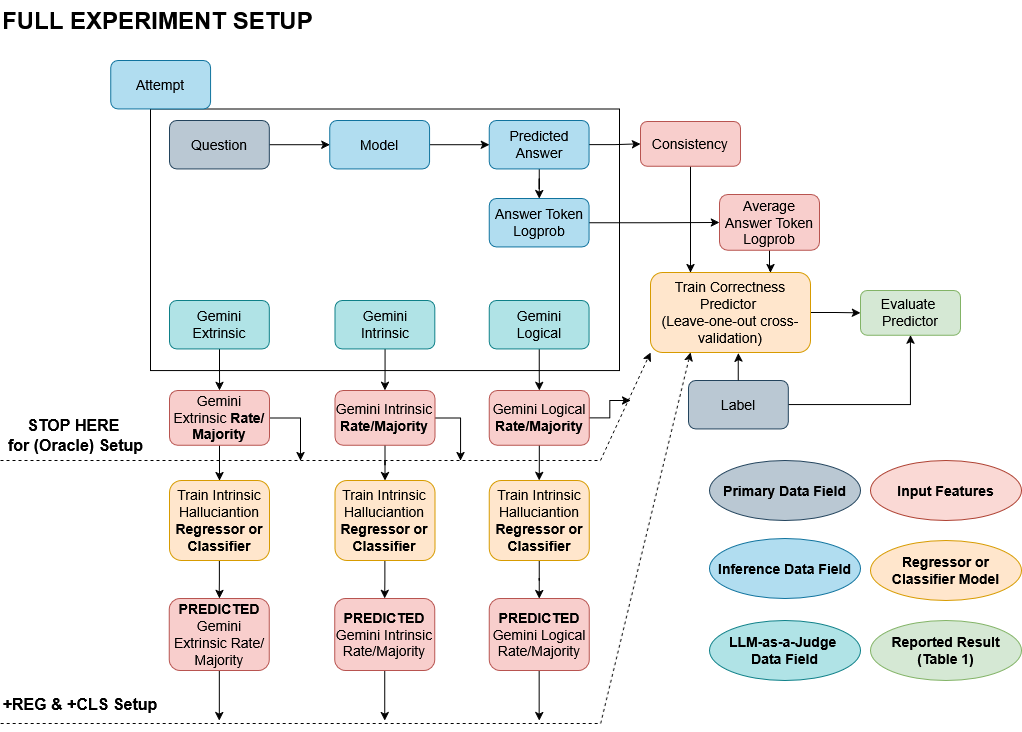, width = 15cm}}
  \caption{Experiment Workflow. Correctness Predictor is trained using differing Input Features dependant on the experiment setup. The \textbf{base setup} utilizes only Consistency and Average Answer Token Logprob. The \textbf{Oracle setup} additionally uses Gemini-2.5-flash's LLM-as-a-Judge results to add three new features to the correctness predictor. The \textbf{+REG and +CLS setup} trains a model to \textbf{predict} Gemini-2.5-flash's LLM-as-a-Judge result and uses these predicted features to train the correctness predictor.}
  \label{fig:app_workflow}
\end{figure*}

Refer to Figure \ref{fig:app_workflow} for the experiment workflow. The performance of the hallucination predictor model can be seen in Table \ref{tab:halluc-predictors}.

\subsection*{Hallucination Extraction Prompt}
\label{app:1_haluprompt}
We use the following prompt to extract the judgement from Gemini-2.5-flash.

\begin{lstlisting}[
    basicstyle=\sffamily\small,
    columns=fullflexible,       
    breaklines=true,
    frame=tb,         
    mathescape=false
]
You are a meticulous, expert evaluator in the field of prosthodontics. Your task is to perform a three-part evaluation of the "Candidate Explanation" using the "Ground Truth Context" and the detailed "Evaluation Rubric" below.

**Evaluation Rubric:**

1.  **Intrinsic Hallucination:** Does the Candidate Explanation directly CONTRADICT a fact in the Reference Explanation?
    - "Yes" or "No".

2.  **Extrinsic Hallucination:** Does the Candidate Explanation FABRICATE new information not present in the Reference Explanation? Classify any fabrication based on its **harm to the explanation's core argument**:
    - **"Harmful":** The fabricated information is an **essential detail** to support the final answer. It introduces a new cause or reason that is essential to the logic, deviating from the core premise of the ground truth.
    - **"Benign":** The fabricated information is a **superfluous detail**. It does not deviate from the core premise of the ground truth.
    - **"None":** The explanation contains no new information.

3.  **Logical Hallucination:** Does the reasoning in the Candidate Explanation logically support the Candidate's Final Answer?
    - "Aligned" or "Misaligned".

**Your Task:**
Provide your evaluation ONLY in the following JSON format:
{{
  "intrinsic_hallucination": "Yes or No",
  "extrinsic_hallucination": "None, Benign, or Harmful",
  "logical_rating": "Aligned or Misaligned",
  "reasoning": "An explanation for your evaluation, referencing the rubric."
}}

---
**Ground Truth Context:**
**Question:** {question_text}
**Reference Explanation:**
{ground_truth_explanation}
---
**Candidate to Evaluate:**
**Candidate's Final Answer:** {prediction_answer}
**Candidate Explanation:**
{prediction_explanation}
---
**JSON Output:**
\end{lstlisting}

\end{document}